\documentclass[10pt,twocolumn,letterpaper]{article}

\usepackage{cvpr}
\usepackage{times}
\usepackage{epsfig}
\usepackage{graphicx}
\usepackage{amsmath}
\usepackage{amssymb}

\usepackage[pagebackref=true,breaklinks=true,letterpaper=true,colorlinks,bookmarks=false]{hyperref}
\cvprfinalcopy 

\def\cvprPaperID{6096} 

\usepackage[utf8]{inputenc} 
\usepackage[T1]{fontenc}    
\usepackage{times}
\usepackage{epsfig}
\usepackage{graphicx}
\usepackage{colortbl}
\definecolor{mygray}{gray}{.92}
\definecolor{gray}{gray}{.5}

\usepackage{amsmath,amsfonts,bm}

\def\eqref#1{equation~\ref{#1}}

\def\1{\bm{1}}

\DeclareMathAlphabet{\mathsfit}{\encodingdefault}{\sfdefault}{m}{sl}
\SetMathAlphabet{\mathsfit}{bold}{\encodingdefault}{\sfdefault}{bx}{n}

\def\sD{{\mathbb{D}}}

\usepackage[pagebackref=true,breaklinks=true,letterpaper=true,colorlinks,bookmarks=false]{hyperref}
\usepackage[utf8]{inputenc} 
\usepackage[T1]{fontenc}    
\usepackage{hyperref}       
\usepackage{url}            
\usepackage{booktabs}       
\usepackage{amsfonts}       
\usepackage{amsmath}
\usepackage{amssymb}
\usepackage{nicefrac}       
\usepackage{microtype}      
\usepackage{adjustbox}
\usepackage{booktabs}
\usepackage{multirow}
\usepackage{adjustbox}
\usepackage{xcolor}
\usepackage{graphics,psfrag}
\usepackage{diagbox} 
\usepackage{subfigure}
\usepackage{bbding}
\usepackage{amssymb}
\usepackage{times}
\usepackage{epsfig}
\usepackage{algorithmic}
\usepackage{algorithm2e}
\usepackage{wrapfig}
\usepackage{float}
\usepackage{rotating}
\usepackage{makecell}
\usepackage{pifont}

\definecolor{ForestGreen}{rgb}{0.13, 0.55, 0.13}
\definecolor{Maroon}{rgb}{0.69, 0.19, 0.0}
\newcommand{\cmark}{\ding{51}}%
\newcolumntype{C}[1]{>{\centering\arraybackslash}m{#1}}
\newcolumntype{R}[1]{>{\raggedleft\arraybackslash}m{#1}}
\newcolumntype{P}[1]{>{\raggedright\arraybackslash}p{#1}}
\newcolumntype{M}[1]{>{\centering\arraybackslash}m{#1}}
\def\etal{\emph{et al.}}

\makeatletter
\ifcvprfinal\fi

\begin{document}
	\pagenumbering{gobble} 
	\title{Multi-Target Domain Adaptation with Collaborative Consistency Learning}
	
	\def\@fnsymbol#1{\ensuremath{\ifcase#1\or \dagger\or* \or \ddagger\or
			\mathsection\or \mathparagraph\or \|\or **\or \dagger\dagger
			\or \ddagger\ddagger \else\@ctrerr\fi}}
	
	\author{%
		Takashi Isobe$^{1,3}$\thanks{The work was done in Noah's Ark Lab, Huawei Technologies.}, Xu Jia$^2$$^*$, Shuaijun Chen$^3$, Jianzhong He$^3$, Yongjie Shi$^4$, \\ Jianzhuang Liu$^3$, Huchuan Lu$^2$, Shengjin Wang$^1$\thanks{Corresponding author}\\
		{$^1$Department of Electronic Engineering, Tsinghua University}\\
		{$^2$Dalian University of Technology}\\
		{$^3$Noah's Ark Lab, Huawei Technologies}\\
		{$^4$Key Laboratory of Machine Perception (MOE), Peking University}\\
		{\texttt{\small{jbj18@mails.tsinghua.edu.cn}} \hspace{0.5cm}}
		{\texttt{\small{wgsg}@tsinghua.edu.cn}\hspace{0.5cm}} \\
		{\texttt{\small{shiyongjie}@pku.edu.cn}\hspace{0.5cm}}
		{\texttt{\small{\{xjia, lhchuana\}}@dlut.edu.cn}\hspace{0.5cm}} \\
		{\texttt{\small{\{chenshuaijun, jianzhong.he, liu.jianzhuang\}@huawei.com}}}\\ 
	}

	\maketitle
	
\begin{abstract}
	Recently unsupervised domain adaptation for the semantic segmentation task has become more and more popular due to high-cost of pixel-level annotation on real-world images. However, most domain adaptation methods are only restricted to single-source-single-target pair, and can not be directly extended to multiple target domains. In this work, we propose a collaborative learning framework to achieve unsupervised multi-target domain adaptation. An unsupervised domain adaptation expert model is first trained for each source-target pair and is further encouraged to collaborate with each other through a bridge built between different target domains. These expert models are further improved by adding the regularization of making the consistent pixel-wise prediction for each sample with the same structured context. To obtain a single model that works across multiple target domains, we propose to simultaneously learn a student model which is trained to not only imitate the output of each expert on the corresponding target domain, but also to pull different expert close to each other with regularization on their weights. Extensive experiments demonstrate that the proposed method can effectively exploit rich structured information contained in both labeled source domain and multiple unlabeled target domains. Not only does it perform well across multiple target domains but also performs favorably against state-of-the-art unsupervised domain adaptation methods specially trained on a single source-target pair. Code is available at \url{https://github.com/junpan19/MTDA}.
\end{abstract}
\section{Introduction}
\label{intro}
Semantic segmentation aims at interpreting an image by assigning each pixel to a semantic class~\cite{long2015fully,chen2017rethinking,chen2018encoder,yu2020context,zhu2019asymmetric}. Recently, semantic segmentation has achieved remarkable progress and is widely applied to intelligent systems such as autonomous driving, human-computer interaction and other low-level vision tasks~\cite{isobe2020video,isobe2020video2,isobe2020revisiting}. Its success is mainly attributed to the supervised learning over large amounts of annotated data. However, human efforts on pixel-level annotations are expensive, which substantially limits the scalability of segmentation models. With large amounts of low-cost and diverse synthetic data simulated with game engines available, unsupervised domain adaptation (UDA) draws much attention to adapt the model learned on synthetic data to real-world data. Unsupervised domain adaptation methods~\cite{lee2019sliced,vu2019dada,zhang2019curriculum,luo2019significance,chen2019progressive,zhao2019multi,lv2020cross,mei2020instance} 
alleviate the issue of domain mismatch by training a model on both labeled source domain and unlabeled target domain.

\begin{figure}[t]
	\begin{center}
		\includegraphics[width=1.0\linewidth]{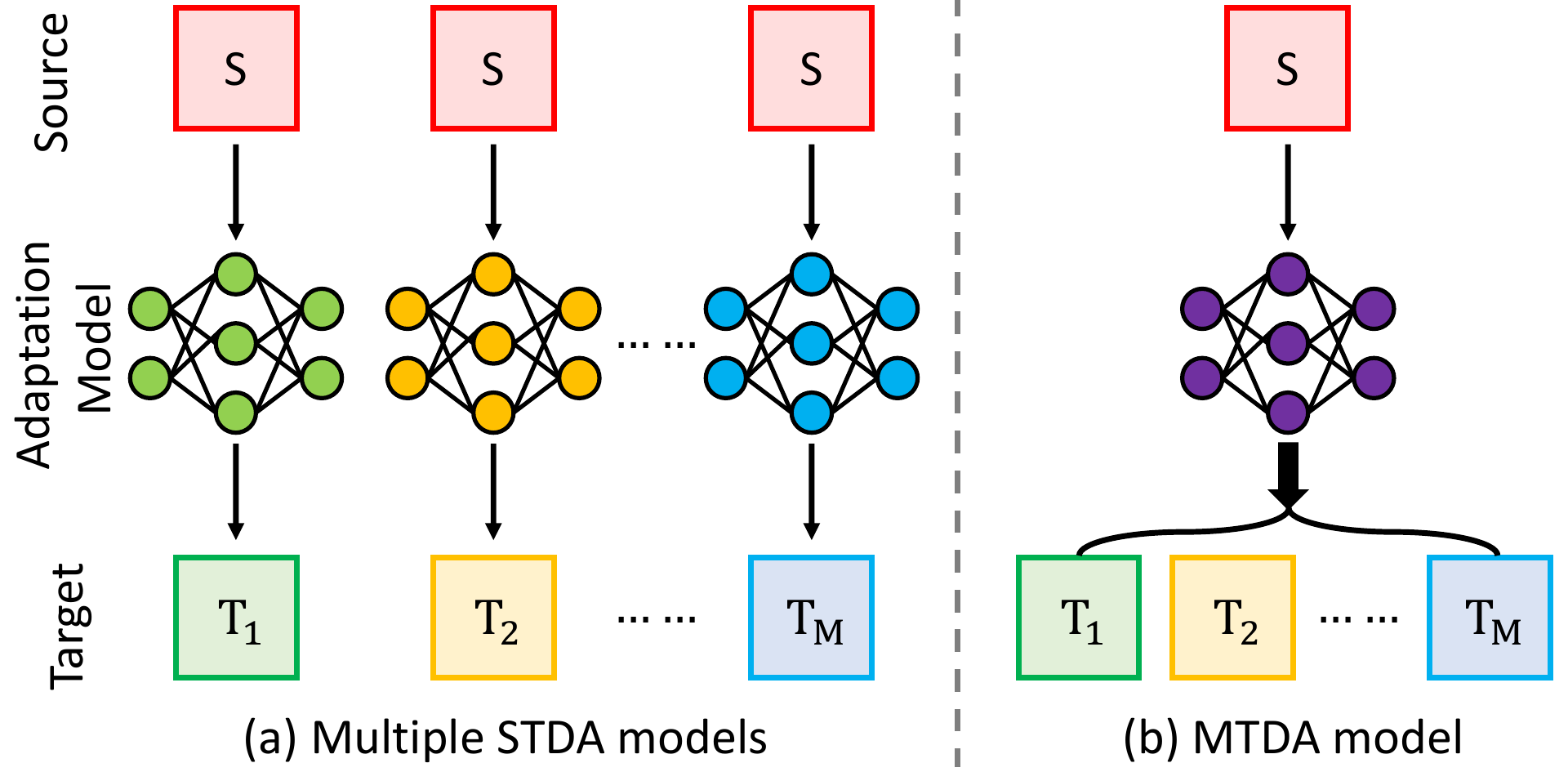}
	\end{center}
	\label{fig:intro}
	\caption{Comparison between the setting of single-target domain adaptation (STDA) and multi-target domain adaptation (MTDA). (a) Multiple STDA models with each one corresponding to a single target domain. (b) A single MTDA model working across multiple target domains.}
	\vspace{-3mm}
\end{figure}

However, the setting of traditional unsupervised domain adaptation in semantic segmentation is usually restricted to single-source-single-target pair, as shown in Figure~\ref{fig:intro} (a). The learned model only works for a single target domain and can not be easily extended to multiple target domains, that is, multi-target domain adaptation (MTDA). With this setting, it is expected to learn a single model that is able to make full use of data from a single labeled source domain and multiple unlabeled target domains and performs well on multiple target domains simultaneously. This setting has great value in real-world applications. For example, in autonomous driving it is expected to have a model work in various environments with different lighting, weather and cityscapes. It is difficult to collect annotated data for such different environments but is easy to have large amounts of unlabeled data.

There have been several works on MTDA~\cite{gholami2020unsupervised,nguyen2020unsupervised,yu2018multi}, however, most of them focus on the classification task. Few works are developed to address the semantic segmentation task under the setting of multi-target domain adaptation. To the best of our knowledge, this is the first work to explore multi-target domain adaptation for semantic segmentation. The main challenge with this task are two folds: (1) lack of pixel-wise supervised information in multiple target domains poses great difficulty in mining inherent and transferable knowledge; (2) it is difficult to have a single model that works well on multiple target domains. 
There are two intuitive ways of extending the pair-wise DA to work on multiple target domains: (1) training multiple models individually for each target domain and (2) training a single model on combined data from multiple target domains. However, directly using multiple models would not play the model ensembling effect as in that in single domain. Inaccurate model dispatching would increase the risk of danger in practical applications. The model developed by direct data combination is likely to incur performance degradation due to the discrepancy between domains.  Intuitively, a generic expert learned in a naive way might have inferior knowledge than the specialized expert for each target domain.

In this paper, we propose a novel collaborative consistency learning framework for multi-target domain adaptation, which includes collaborative consistency learning among multiple expert models and online knowledge distillation to obtain a single domain-generic student model. This work shows that once connection among domains is fully explored, \textit{i.e.}, connection between each source-target domain pair and among target domains, it can obtain even better performance than models learned with unsupervised domain adaptation methods for each source-target domain pair. 

In the proposed collaborative consistency learning framework, data from all domains are first translated to the style of each target domain, respectively. In this way, we build a bridge between each pair of target domains, that is, images from the same domain are translated into different styles corresponding to different target domains. For each style, a semantic segmentation model is trained on both translated labeled data from source domain and translated unlabeled data from multiple target domains. Each network is a domain-specific expert and is trained with a kind of UDA loss and an additional consistency loss that align segmentation results of images of the same content but with different styles based on the bridge. Such collaborative consistency learning helps knowledge exchange among domain-specific experts.
To obtain a single model that works across multiple target domains, we design a student model whose weights are regularized by the weights of multiple experts and further teach it with multiple experts through knowledge distillation. In this way, the student model is able to learn common semantic knowledge from teachers across multiple domains.


To sum up, we make the following contributions: 
\begin{itemize}
	\item To the best of our knowledge, this is the first work that explores the unsupervised multi-target domain adaptation task in semantic segmentation. 
	\item We propose a new collaborative consistency learning framework to handle the MTDA task for semantic segmentation, where unlabeled data in multiple target domains is fully leveraged to train a single model that works across all target domains.
	\item Experimental results demonstrate the effectiveness of the proposed method. We can obtain a single model that not only works well across multiple target domains but also performs favorably against domain-specialized models on each target domain.
\end{itemize}

\section{Related Work}
\label{related}
\subsection{Unsupervised Domain Adaptation for Semantic Segmentation}
\textbf{Single-target Domain Adaptation.}
A typical practice for UDA in segmentation is to apply a model that is trained on a synthetic source domain to a real target domain. Unfortunately, the domain shift between the synthetic and real data would deteriorate the performance of model generalization~\cite{tsai2018learning,zou2018unsupervised,yang2020label}. There are three main categories of methods to seek a bridge the gap between the source and target domain. The first category is adversarial-based UDA~\cite{tsai2018learning,luo2019taking,chen2018road,li2020content,hoffman2016fcns,huang2020contextual,vu2019advent,pan2020unsupervised} approaches which reduce domain discrepancy by maximizing the confusion between source and target in the feature~\cite{tsai2018learning,luo2019taking,chen2018road,hoffman2016fcns,huang2020contextual} or entropy space~\cite{vu2019advent,pan2020unsupervised}. The second category of methods attempt to learn domain-invariant representation by taking advantage of various image translation techniques~\cite{zhu2017unpaired,huang2017arbitrary}, e.g. target-to-source translation in~\cite{yang2020label}, bidirectional translation in~\cite{li2019bidirectional} and texture-diversified translation in~\cite{kim2020learning}.  
The third category of methods attempt to apply self-training~\cite{zou2018unsupervised,lian2019constructing,li2019bidirectional,li2020content,wang2020differential,kim2020learning,pan2020unsupervised} or model ensembling~\cite{yang2020fda,vu2019advent,chen2019domain} for further improvement in the unlabeled target domain. 
Despite UDA for segmentation is a broadly studied topic, most of the previous works address address the UDA task under the setting of single-target domain adaptation (STDA), which has limitation in practical applications. Moreover, most of the previous works for STDA focus on fully utilizing the labeled data to improve the performance in unlabeled domain~\cite{huang2020contextual,chang2019all,yang2020label}. We argue that fully utilize the unlabeled data is also beneficial to explore the informative information within unlabeled data, thus improve the final performance on target domain. Based on these observations, multi-target domain adaptation (MTDA) is more realistic setting in real-world.  

\textbf{Multi-target Domain Adaptation.}
There are two naive ways of directly extending domain-specialized UDA to work on multiple target domains, that are (1) training multiple models individually for each target domain (2) training a single model on combined data from multiple target domains. Unfortunately, these methods are not appropriate to handle MTDA problem because they would suffer from performance degradation due to the mismatching of multi-target domains. Despite several works have been done to address the MTDA task, they just focus on addressing classification task~\cite{gholami2020unsupervised,nguyen2020unsupervised,yu2018multi}. MTDA for segmentation is more challenging as it is in essence a dense pixel prediction task. The work most related to ours is~\cite{nguyen2020unsupervised}, which also applies multiple teachers to obtain a common knowledge model for each target domain. However, in~\cite{nguyen2020unsupervised}, unlabeled data from different target domains are not fully exploited to train stronger teachers and there is not any regularization in online knowledge distillation on both the student and teachers. 

\textbf{Domain Generalization.}
The task of MTDA is also related to Domain generalization (DG), which attempts to generalize a model trained only on source domain to multiple unseen target domains by learning domain-invariant feature of source~\cite{khosla2012undoing,dou2019domain,balaji2018metareg,zhang2020generalizable,li2018deep,yue2019domain}. Khosla~\etal~\cite{khosla2012undoing} proposed removing the data bias by factoring out the domain-specific and domain-agnostic component during training on source domains. Yue~\etal~\cite{li2018deep} proposed learning a domain-invariant feature representation via adversarial training. In~\cite{yue2019domain}, domain randomization and consistency-enforced training are both used to learn a domain-invariant network with synthetic images. Compared to the task of DG, where data from target domain is absent, the MTDA task aims at training a model for multiple target domains by fully exploring the unlabeled data. 

%
\subsection{Knowledge Distillation}
Knowledge distillation (KD) has been widely studied for learning a compacting and fasting model for edge devices in real-world applications including face recognition, super-resolution and object detection. The idea of KD is first proposed by~\cite{hinton2015distilling}, in which a student model is used to mimic the distribution of teacher's prediction. By transferring the knowledge from teacher to student, the student model is on par with or even better performance than the teacher model~\cite{furlanello2018born,mirzadeh2020improved,heo2019comprehensive,noroozi2018boosting,kang2020towards}. Rather than training a student to distill knowledge from a pretrained teacher, Zhang~\etal~\cite{zhang2018deep} proposed to learn an ensemble of students which collaboratively teach each other throughout the training process.
In this paper, we share similar philosophy as the general KD and adapt it to the MTDA task. Multiple domain-specific expert models with promising performance in each target domain are adopted as teacher, and a student is expected to perform well across all target domains. The student is taught simultaneously by multiple teachers, and also gives feedback to all teachers, all of which are implemented in an online fashion. 
gives rise to robust domain-invariant CNNs trained
using synthetic images.
%


\section{Methodology}
\label{method}

\subsection{Overview}
\begin{figure*}[t]
	\begin{center}
		\includegraphics[width=1.0\linewidth]{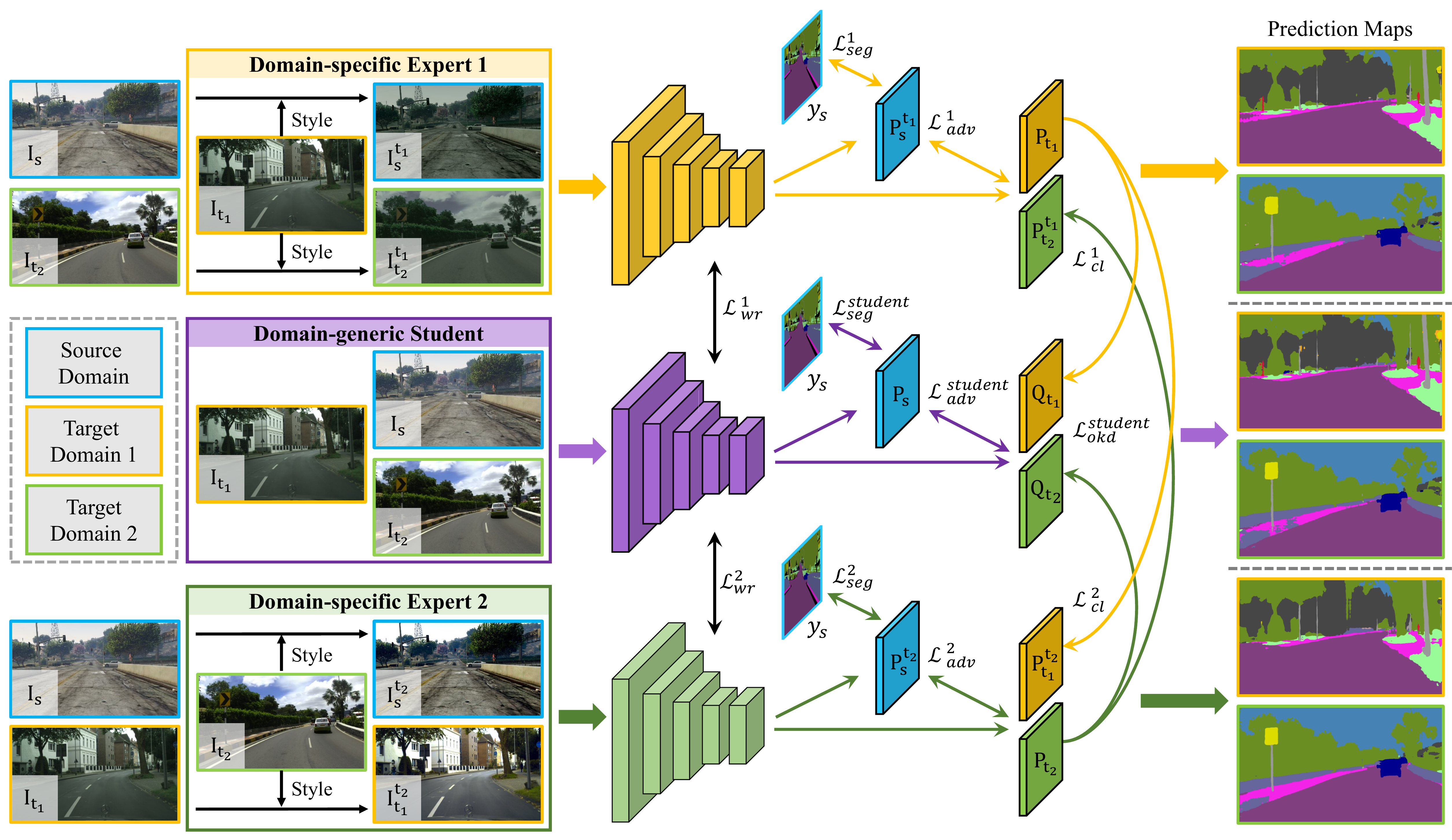}
	\end{center}
	\caption{Overview of the proposed Collaborative Consistency Learning (CCL) framework for MTDA in semantic segmentation. The framework is illustrated with $M=2$ as example but it also holds for other numbers of target domains. Blue, yellow and green box represents the source, the 1-st and the 2-nd target domains, respectively. }
	\label{fig:framework}
	\vspace{-3mm}
\end{figure*}

We propose a novel framework to tackle the task of MTDA for semantic segmentation. Since only images from source domain have annotation maps, the key to this task is to make full use of given source domain data and to explore the way of mining rich structured information contained in unlabeled target domains. Our solution is to first train an expert model for each target domain, which is further encouraged to collaborate with each other simultaneously through a bridge built among different target domains. Since our final goal is to obtain a single model that works well on all target domains, we take the above expert models as teachers and additionally train a student model. It learns not only to imitate the output of each expert on the corresponding target domain but also to pulls different expert close to each other with regularization on their weights. The overall framework is illustrated in Figure~\ref{fig:framework}. Note that all these are done in parallel at the same time. 

Formally, we denote data from source domain as $\sD_s=\{(I_s, y_s)\}$ and data from the $m$-th target domain as $\sD_{t_m}=\{I_{t_m}\}$, where $I_s$ and $y_s$ represent images and the associated pixel-wise annotation. The goal of our work is to adapt the knowledge from $\sD_s$ to $M$ target domains ${\sD_{t_m}}$ which are not associated with any annotation map.
\subsection{Collaborative Consistency Learning for MTDA}

\textbf{Learning of multi-target domain experts.}
For each source-target domain pair, we train a domain adaptation model with most existing unsupervised domain adaptation method~\cite{vu2019advent,tsai2018learning}. In this work, we train a model with a combination of cross-entropy loss on source domain $\sD_s$ for segmentation and adversarial loss for structure adapting, similar to~\cite{vu2019advent,tsai2018learning}. However, instead of directly learning an expert with only data from each source-target pair, the proposed method would learn an expert with data available from all domains. Specifically, as for an expert of a particular target domain, style transfer method is first applied to translate data from all domains to the style of that target domain. In this way, discrepancy between different domains is reduced to some extent. With different semantic contexts but the same style helps learning a UDA expert model for a particular domain. In addition, re-styled data also works as a bridge to connect different target domains for knowledge exchange.
The expert model for the $m$-th target domain is jointly optimized with supervised segmentation loss $\mathcal{L}^m_{seg}$ and adversarial loss $\mathcal{L}^m_{adv}$ as follows:
\begin{equation}
	\begin{aligned}
	\mathcal{L}^{m} = {} &
    \mathcal{L}^m_{seg}(P_s^{t_m}, y_s) + \lambda_{adv}~\mathcal{L}^m_{adv},
	\end{aligned}
\end{equation}
where $P$ is the output of the last layer of domain-specific expert.
For $I^{(\cdot)}_{(\cdot)}$ and $P^{(\cdot)}_{(\cdot)}$, superscript represents the translated style and subscript represents the corresponding domain. $\mathcal{L}^m_{seg}$ indicates the cross-entropy objective between the probability map and its pixel-level annotation map $y_s$. $\lambda_{adv}$ controls the weight of adversarial loss. 
$\mathcal{L}^{m}_{adv}$ is defined as:
\begin{equation}
\begin{aligned}
	\mathcal{L}^m_{adv} = {} &
	 \mathop{\mathbb{E}}[log(1-D^m(P_{t_m}))] + \mathop{\mathbb{E}}[log{D^m(P^{t_m}_s)}]  
	\\ & + \sum\limits_{\substack{n=1 \\ n\neq m}}^M \mathop{\mathbb{E}}[log(1-D^m(P^{t_m}_{t_n}))] + \mathop{\mathbb{E}}[log{D^m(P^{t_m}_s)}],
\end{aligned}
\end{equation}
which enforces the model to align multiple target domains with source domain and learn domain-invariant information with adversarial training. $D^m$ is a discriminator to classify the probability map whether from the source or the integrated target domain which is composed of multiple translated target domains. Note that all experts share the same network architecture but each one has a different set of weights.

\textbf{Knowledge exchange with collaborative consistency learning.}
The above expert domain adaptation models are able to give a reasonable performance on the corresponding domain adaptation task. However, power within data from multiple unlabeled target domains has not been fully exploited. 
As for data from a certain target domain, it has been translated into different styles of other target domains but with the same semantic context reserved. Multiple expert models are trained to make the consistent pixel-wise prediction for each sample with the same semantic context. Since different expert models are learned on samples of different styles, they learn the pixel-wise classification ability in different ways, and their predictions vary from each other. It is such different predictions that provide an opportunity to learn complementary knowledge from other experts and extract essential information that really matters to the performance of semantic segmentation. Therefore, we exploit collaborative learning for knowledge exchange among multiple expert models.
The knowledge exchange with collaborative learning from other experts to the $m$-th expert can be formulated as:  
\begin{equation}
\begin{aligned}
\mathcal{L}^{m}_{cl} = \frac{1}{M-1} \sum\limits_{\substack{n=1 \\ n\neq m}}^M \mathcal{D}_{KL}(P_{t_n}||P_{t_n}^{t_m}), 
\end{aligned}
\end{equation}
where $\mathcal{D}_{KL}$ is average of Kullback-Leibler (KL)-divergence between the probability map $P_{t_n}^{t_m}$ and $P_{t_n}$. The expert of the domain $m$ is trained to imitate the output distribution of other $M$-1 domain experts by $\mathcal{L}_{cl}$. Such knowledge exchange encourages each expert to make full use of unlabeled data in an unsupervised manner. 
The overall objective function of the $m$-th domain-specific expert is optimized by:
\begin{equation}
\mathcal{L}^{expert} = \frac{1}{M} \sum\limits_{\substack{n=1}}^M (\mathcal{L}^{n} + \lambda_{cl}~\mathcal{L}^{n}_{cl}),
\end{equation}
where $\lambda_{cl}$ leverages the importance of consistency loss.
\subsection{Online Knowledge Distillation from Multiple Experts}
We have explained how to train multiple domain-specialized experts by making full use of available labeled and unlabeled data to improve their capability. However, our final purpose is to obtain a single model that performs well across multiple target domains. We propose to online distill knowledge from multiple expert models with additional regularization on their model weights. Specifically, a student network is added to the framework and is supervised with the output of multiple experts. 
\begin{equation}
\mathcal{L}^{student}_{okd} = \frac{1}{M} \sum\limits_{\substack{n=1}}^M \mathcal{D}_{KL}(P_{t_n}||Q_{t_n}),
\end{equation}
where $Q$ is the output of the last layer of the domain-generic student. Then, the overall optimization objective of domain-generic student model can be defined as:

 \begin{equation}
 \small
 \begin{aligned}
 \mathcal{L}^{student} = {} 
 \mathcal{L}_{seg}^{student}(Q_s, y_s) + \lambda_{adv}\mathcal{L}^{student}_{adv} 
  + \lambda_{okd}\mathcal{L}^{student}_{okd} ,
  \label{eq:domain-exp}
 \end{aligned}
 \end{equation}
where $\lambda_{okd}$ is the weight factor to balance the training of online knowledge distillation and weights regularization, respectively. $\mathcal{L}_{seg}^{student}$ means the cross-entropy objective function between the probability map $Q_s$ and its pixel-level annotation map $y_s$. The adversarial loss $\mathcal{L}^{student}_{adv}$ is expressed as:  

\begin{equation}
\begin{aligned}
\mathcal{L}^{student}_{adv}= {}&
 \frac{1}{M} \sum\limits_{\substack{n=1}}^M \mathop{\mathbb{E}}[log(1-D^{student}(Q_{t_n}))]\\& + \mathop{\mathbb{E}}[log{D^{student}(Q_s)}], 
 \end{aligned}
\end{equation}
where $D^{student}$ is a discriminator for training domain-generic student model. However, the performance of directly forcing a student to learn from multiple experts is limited due to diversity among multiple experts. The student might get confused in simultaneously distilling knowledge from very different experts. To address this issue, we propose to pull domain-specific experts a bit closer to the student. In this way, the gap between experts is reduced and it is easier for the student to distill common useful knowledge from these experts.
The gap between domain-specific experts $\{F^m_{expert}\}_{m=1}^M$ and domain-generic student $F_{student}$ can be reduced with the following the weights regularization term:
\begin{equation}
\mathcal{L}_{wr} = \frac{1}{M} \sum\limits_{m=1}^M ||\theta^m - \theta^{student}||_1, 
\end{equation}
where $\theta^m$ and $\theta^s$ represents the weights of the $m$-th domain-specific expert model and the domain-generic student model, respectively. 
The overall optimization objective of the CCL framework can be defined as:
 \begin{equation}
\mathcal{L} = \mathcal{L}^{student} + \mathcal{L}^{expert} + \lambda_{wr}\mathcal{L}_{wr},  
\end{equation}
where $\lambda_{wr}$ is the weighting parameters.
Finally, the obtained domain-generic model is applied across $M$ target domains. 

\section{Experiments}
\label{experiment}
In this section, we describe the experiment setting and implementation details of the proposed CCL. Extensive ablation studies and comparison with other MTDA and STDA methods are also provided. We show that our method can work well on multiple large scale urban driving datasets.
\begin{table*}[t]
	\footnotesize
	\renewcommand\arraystretch{1.2}
	\setlength\tabcolsep{2.5pt}
	\caption{Performance comparison between our method and baseline models on adaptation from GTA5 to Cityscapes and IDD. The mIoU is calculated by the average of the intersection-over-union (IoU) among all 19 categories. "R" represents the ResNet101-based model and "V" represents the VGG16-based model. "C" and "I" indicate the target domain on Cityscapes and IDD, respectively. "*" represents the method with multiple models that are individually trained for each target domain.}
	\vspace{-3pt}
	\begin{center}
		\begin{tabular}{ l| c|c|c c c c c c c c c c c c c c c c c c c| c}
			\toprule
			\multicolumn{23}{ c }{\bf GTA5 $\rightarrow$ Cityscapes \& IDD } \\
			\hline	
			Method &\rotatebox{90}{Model} &\rotatebox{90}{Target} &\rotatebox{90}{road} &\rotatebox{90}{sidewalk} &\rotatebox{90}{building} &\rotatebox{90}{wall} &\rotatebox{90}{fence} &\rotatebox{90}{pole} &\rotatebox{90}{light} &\rotatebox{90}{sign} &\rotatebox{90}{veg.} &\rotatebox{90}{terrain} &\rotatebox{90}{sky} &\rotatebox{90}{person} &\rotatebox{90}{rider} &\rotatebox{90}{car} &\rotatebox{90}{truck} &\rotatebox{90}{bus} &\rotatebox{90}{train} &\rotatebox{90}{motor} &\rotatebox{90}{bike} &{\bf{mIoU}} \\ 
			\hline
			\multirow{2}{*}{Individual Model*} &\multirow{2}{*}{V} & C
			&88.4  &30.8  &78.4  &29.8  &25.9  &20.5  &17.6  &11.2  &79.2  &30.3  &65.1  &46.6  &9.1  &81.2  &22.9  &29.9  &0.1  &11.9  &0.5  &35.8 \\
			
			&  &I &68.8  &2.5  &61.4  &29.2  &20.8  &24.9  &7.3  &34.3  &75.6  &29.3  &91.2  &39.8  &28.3  &63.6 &35.8  &38.8  &0  &39.2  &7.8  &36.8  \\
			\hline
			
			\multirow{2}{*}{Source only} &\multirow{2}{*}{V} & C
			&64.0  &16.8  &67.0  &22.6  &18.9  &22.1  &20.6  &13.3 &76.8  &14.8  &63.9  &47.9  &5.7  &72.5  &12.3  &12.9  &9.5  &19.1  &2.3  &30.7 \\
			
			&  &I &50.9  &2.3  &45.8  &21.8  &20.5  &26.8  &6.8  &39.6  &76.1  &28.3  &82.0  &38.6  &28.8  &69.2  &38.2  &16.6  &0  &49.1  &9.7  &34.3  \\
			\hline		
			\multirow{2}{*}{Data Combination} &\multirow{2}{*}{V}  & C
			&86.8  &16.1  &77.1  &27.8  &16.6  &22.1  &16.4  &6.1  &80.9  &30.9  &68.0  &43.2  &8.9  &80.7  &23.3  &15.2  &0  &11.0  &1.3  &33.3   \\
			
			&  &I &73.8  &3.5  &52.3  &25.8  &19.4  &24.6  &8.4  &32.0  &78.9  &32.2  &84.6  &38.6  &37.5  &73.1  &38.5  &12.9  &0  &41.3  &5.1  &35.9  \\
			\hline	
			\multirow{2}{*}{Ours}  &\multirow{2}{*}{V}  & C
			&89.3  &33.6  &79.6  &26.8  &22.6  &25.9 &25.1  &17.7  &81.8  &32.9  &72.3  &49.4  &15.2  &82.0  &22.5  &16.9  &9.6  &10.7  &4.3  &\bf 37.8 \\
			
			&  &I &85.4  &5.8  &64.2  &31.8  &19.2  &24.9  &5.6  &43.2  &77.3  &35.04  &91.3  &43.9  &37.6  &70.1  &42.2  &27.5  &0  &46.9  &9.7  &\bf 40.1 \\	
			\hline
			\hline
			\multirow{2}{*}{Individual Model*} &\multirow{2}{*}{R} & C
			&88.8  &23.8 &81.5 &27.7 &27.3 &31.7  &33.2 &22.9  &83.1 &27.0 &76.4  &58.5  &28.9  &84.3  &30.0  &36.8  &0.3  &27.7  &33.1  &43.3  \\
			
			&  &I &94.1  &24.4  &66.1  &31.3  &22.0  &25.4  &9.3  &26.7  &80.0  &31.4  &93.5 &48.7  &43.8  &71.4  &49.4  &28.5  &0 &48.7 &34.3 &43.6 \\
			\hline
			\multirow{2}{*}{Source only} &\multirow{2}{*}{R} & C
			&79.0  &9.2  &76.1  &15.7  &17.1  &23.3  &28.0  &14.8  &82.4  &22.9  &70.8  &53.7   &27.1  &76.6  &35.9  &5.4  &0.7  &20.3  &39.6  &36.8  \\
			
			& &I &60.5  &8.3  &50.8  &8.2  &18.9  &27.0  &6.2  &33.3  &67.6  &22.4  &87.4  &52.0  &45.8  &71.8  &43.9  &37.1  &0  &50.7  &20.2  &37.5  \\
			\hline	
%
%
			\multirow{2}{*}{Data Combination} &\multirow{2}{*}{R} & C
			&86.1  &32.0  &79.8  &24.3  &22.3  &28.5  &27.9  &14.3  &85.1  &29.8  &79.9  &56.1  &20.5  &77.7  &34.4  &35.2  &0.7  &18.2  &13.1  &40.3 \\
			
			&  &I &92.8  &23.4 &60.9  &25.8  &23.4  &24.1  &8.6  &32.2  &77.5  &26.8  &92.3 &48.0  &41.0  &74.4  &48.4  &17.7  &0 &52.5  &28.2 &42.0  \\
			\cline{1-23}
			
			\multirow{2}{*}{Ours} &\multirow{2}{*}{R} & C
			&90.3  &34.0  &82.5  &26.2  &26.6  &33.6 &35.4  &21.5  &84.7  &39.8  &81.1  &58.4  &25.8  &84.5  &31.4  &45.4  &0  &29.9  &24.7  &\bf45.0 \\
			
			&  &I &95.0  &30.5  &65.6  &29.4  &23.4  &29.2  &12.0  &37.8  &77.3  &31.3  &91.9  &52.4  &48.3  &74.9  &50.1  &36.6  &0  &56.1  &32.4  &\bf46.0 \\
			\hline	
		\end{tabular}
		\label{table:gta2city+iDD}
	\end{center}
\end{table*}

%

\begin{table}[t]
	\footnotesize
	\setlength\tabcolsep{3.5pt}
	\caption{Comparison of our model with SOTA UDA methods, DG methods and MTDA methods with ResNet-101 as backbone. The mIoU and mIoU* are evaluated over the 19 and 13 classes, respectively. "G", "S", "C" and "I" represent "GTA5", "SYNTHIA", "Cityscapes" and "IDD", respectively. ${\dagger}$ means the results of our implementation. All numbers correspond to the results without using pseudo labels or model ensembling as reported in the original papers.}
	\vspace{-3pt}
	\begin{center}
		\begin{tabular}{P{1cm} | C{2.2cm}  | C{0.9cm}  C{0.9cm} | C{0.9cm}   C{0.9cm} }
			\toprule
			\multirow{2}{*}{Setting} &\multirow{2}{*}{Method}  &\multicolumn{2}{c|}{mIoU} &\multicolumn{2}{c}{mIoU*} \\ 
			& &G $\to$ C  &G $\to$ I &S $\to$ C &S $\to$ I  \\
			\midrule
			\multirow{9}{*}{STDA} &AdaptSeg~\cite{tsai2018learning} &42.4  &-  &46.7  &-      \\
			&CLAN~\cite{luo2019taking}                           &43.2  &-  &47.8  &-     \\
			&ADVENT~\cite{vu2019advent} 	                     &43.8  &-  &47.8  &-     \\
			&BDL~\cite{li2019bidirectional}                      &41.1  &-  &-  &-     \\
			&SIBAN~\cite{luo2019significance}                    &42.6  &-  &46.3  &-     \\
			&AdaptPatch~\cite{tsai2019domain}                    &44.9  &-   &-  &-    \\
			&MaxSquare~\cite{chen2019domain}	                 &44.3  &-  &45.8 &-    \\
			&Kim et al.~\cite{kim2020learning}                   &44.6 &-  &-  &-   			 \\
			&FDA~\cite{yang2020fda}                				 &44.6  &-  &-  &-    \\
			&IntraDA~\cite{pan2020unsupervised}         		 &46.3  &-  &48.9  &-    \\
			\midrule
			DG &Yue et al.$^{\dagger}$~\cite{yue2019domain} &42.1  &42.8  &44.3  &41.2   \\
			\midrule
			\multirow{3}{*}{MTDA}&MTDA-ITA$^{\dagger}$~\cite{gholami2020unsupervised}	&40.3  &41.2  &42.7  &39.4  \\
			&MT-MTDA$^{\dagger}$~\cite{nguyen2020unsupervised} 			 &43.2  &44.0  &45.2  &42.2   \\
			&Ours 			  							 &45.0  &46.0  &48.1  &44.0   \\
			\bottomrule
		\end{tabular}
		\label{table:syn2city+idd-sota}
	\end{center}
	\vspace{-5mm}
\end{table}

\begin{table}[t]
	\footnotesize
	\setlength\tabcolsep{3.5pt}
	\caption{Ablation studies of the proposed CCL framework on GTA5 to Cityscapes and IDD with ResNet-101 as backbone. }
	\vspace{-3pt}
	\begin{center}
		\begin{tabular}{C{1.1cm}| C{1.1cm} C{1.1cm} C{1.1cm} |  C{1.1cm}  C{1.1cm} }
			\toprule
			Model \# &$\mathcal{L}_{cl}$ &$\mathcal{L}_{okd}$ &$\mathcal{L}_{wr}$ &C &I\\
			\midrule
			1& &  &  	&42.3  &42.9  \\
			\midrule
			2&	 &\cmark  & &41.8  &43.9  \\
			3&  & &\cmark 	&43.1  &43.5     \\
			4&  &\cmark  &\cmark 	&44.0  &44.7  \\
			\midrule
			5&\cmark  &\cmark  & 	&42.4  &45.2   \\
			6&\cmark  &  &\cmark 	&44.2  &44.9   \\
			7&\cmark  &\cmark  &\cmark 	&45.0  &46.0  \\
			\midrule
			\multicolumn{4}{c|}{Individual Model} 	&43.3  &43.6  \\
			\bottomrule
		\end{tabular}
		\label{table:ablation_CCL}
	\end{center}
	\vspace{-5mm}
\end{table}

\subsection{Datasets}


Under the MTDA experiment setting, synthetic datasets including GTA5~\cite{richter2016playing} and SYNTHIA~\cite{ros2016synthia} are used as source domain respectively, along with multiple real-world datasets Cityscapes~\cite{cordts2016cityscapes}, Indian Driving (IDD)~\cite{varma2019idd} and Mapillary~\cite{neuhold2017mapillary} as the target domains. The proposed CCL model is trained with labeled source data and unlabeled target data from various domains. Results on the validation sets of the datasets corresponding to the multiple target domains are used to evaluate its performance.

\textbf{GTA5} contains 24,966 synthetic images with a resolution of 1914$\times$1052 pixels that are collected from the video game GTA5 along with pixel-level annotations that are compatible with Cityscapes, IDD and Mapillary in 19 categories.

\textbf{SYNTHIA} is another synthetic dataset. The SYNTHIA-RAND-CITYSCAPES split of SYNTHIA, which contains 9,400 rendered images of 1280$\times$760 resolution, is used as another source domain. We use the 16 common categories with Cityscapes, IDD and Mapillary for training and 13 common classes for testing.

\textbf{Cityscapes} is a real-world dataset with 5,000 street scenes taken from European cities and labeled into 19 classes. We use 2,975 images for training and 500 validation images. 

\textbf{IDD} is a more diverse dataset than Cityscapes which captures unstructured traffic on India’s road. It contains a total of 10,003 images, with 6,993 images for training, 981 for validation and 2,029 for testing.

\textbf{Mapillary} provides 25,000 images collected from all around the world and diverse source of image capturing devices. It includes 18,000 images for training, 5,000 images for testing, and 2,000 images for validation. 

\begin{figure*}[t]
	\begin{center}
		\includegraphics[width=1.0\linewidth]{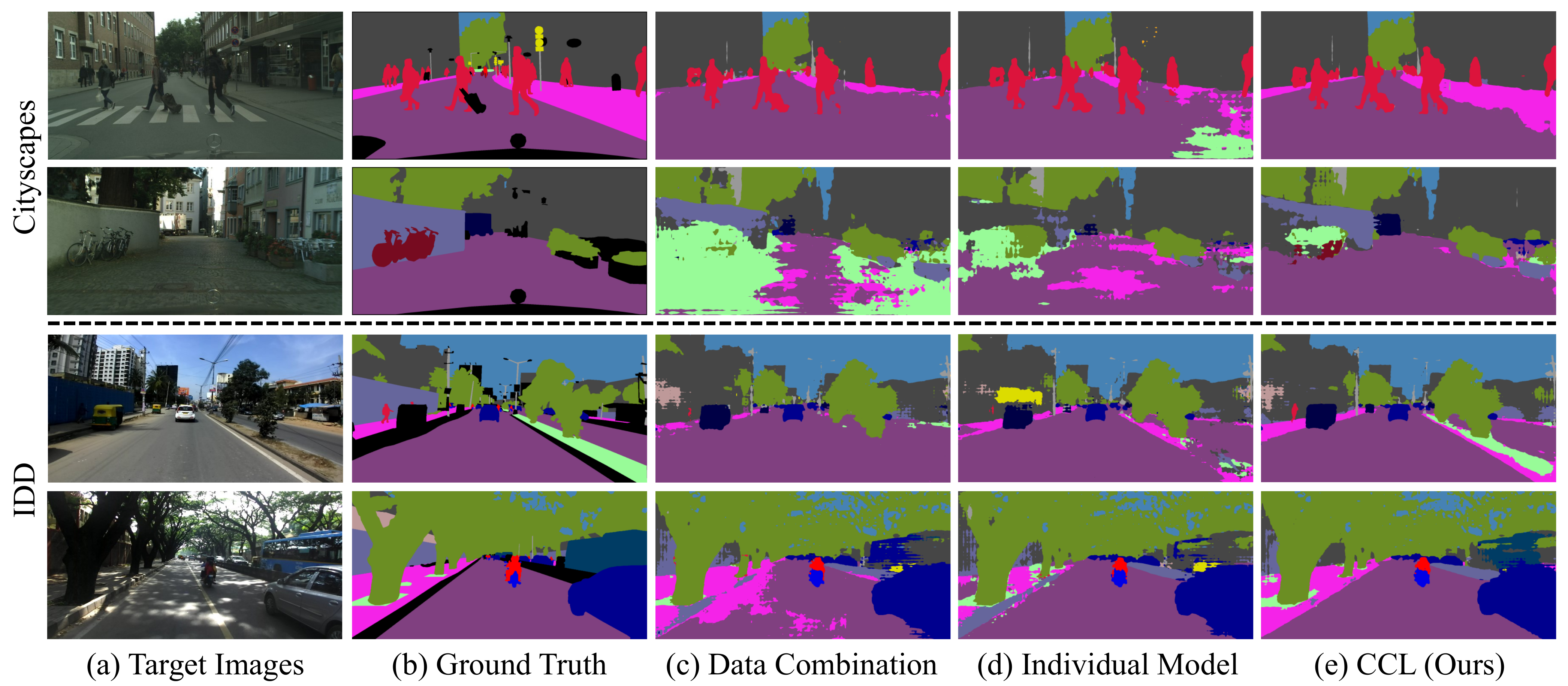}
	\end{center}
	\label{fig:target_semantic}
	\caption{Qualitative results for GTA5 to Cityscapes and IDD.}
	\vspace{-3mm}
	\label{figure:sota}
\end{figure*}

%






\subsection{Training Details}
Similar to~\cite{tsai2018learning} and~\cite{vu2019advent}, we use the DeepLab-v2~\cite{chen2017deeplab} model with ResNet-101~\cite{he2016deep} and VGG-16~\cite{simonyan2014very} as backbones and initialize them with models pre-trained on ImageNet~\cite{deng2009imagenet}. For the discriminator, we also adopt the same network architecture as~\cite{tsai2018learning,vu2019advent}. The semantic segmentation model parameters are optimized with SGD optimizer~\cite{bottou2010large} where the weight decay and momentum are set
to 0.9 and $5\times10^{-4}$, respectively. The learning rate is initially set to $2.5 \times 10^{-4}$. The polynomial procedure~\cite{chen2017deeplab} is used as the learning rate schedule. The discriminator is optimized with Adam optimizer~\cite{kingma2014adam} with the momentum $0.9$ and $0.99$ with the learning rate is set to $10^{-4}$. We set $\lambda_{adv}$, $\lambda_{cl}$, $\lambda_{okd}$ and $\lambda_{wr}$ as $10^{-3}$. Here, we adopt a simple way to conduct image translation in gamut of LAB color space~\cite{reinhard2001color}.

\subsection{Comparison with Baseline Models}
We compare the segmentation performance of the proposed CCL with three baselines: "Individual Model", "Source Only" and "Data Combination". "Individual Model", similar to~\cite{vu2019advent}, is to train multiple models for each corresponding target. "Source Only" and "Data Combination" are the MTDA setting which trains a single model across multiple target domains. "Source Only" is to train a model with the data only from source domain. "Data combination" is trained by directly combine data from multiple target domains as one domain. Here, we conduct the experiment with two target domains (\textit{i.e.,}~$M$=2), but our method can be easily extended to the case of more number of target domains. The results of each method are reported in Table~\ref{table:gta2city+iDD}. In~Table~\ref{table:gta2city+iDD}, the method of "Individual Model" that trains two models individually on Cityscapes and IDD achieves 43.3\% and 43.6\% mIoU on the corresponding domain. However, it requires two models for each domain. Compared to that, "Source only" use a single model but suffers considerable performance drops by 6.5\% and 6.1\% on Cityscapes and IDD because of the domain shift between the synthetic and real data. By directly combining the multiple target data as one domain, the model trained by "Data Combination" also suffers the performance degradation lagging behind the method of "Individual Model" by 3.0\% and 1.6\% mIoU on Cityscapes and IDD. Our method with a single model achieves 45.0\% and 46.0\% mIoU on Cityscapes and IDD, which significantly outperforms the "Data Combination" by +4.7\% and +4.0\%. By fully exploring unlabeled data from multiple target domains, the proposed CCL even works better than the "Individual Model", which adopts two models and trained on each target domain individually, by +1.7\% and +2.4\% mIoU on Cityscapes and IDD. The qualitative comparison between different baselines and the proposed CCL are provided in Figure~\ref{figure:sota}.

\subsection{Comparison with State-of-the-arts}
We first compare our method with the single-target domain adaptation (STDA) method on GTA5-to-Cityscapes and SYNTHIA-to-Cityscapes with using ResNet-101 as backbone. The results are shown in Table~\ref{table:syn2city+idd-sota}. Our method performs favorably against state-of-the-art domain-specialized UDA methods on both GTA5-to-Cityscapes and SYNTHIA-to-Cityscapes. However, it is noteworthy that with one round of training the proposed obtains a single model that achieves good performance on both Cityscapes and IDD. We also compare our method with DG and MTDA on "GTA5 to Cityscapes and IDD" and "SYNTHIA to Cityscapes and IDD". Compared to the method of DG, where the unlabeled data were not be used in~\cite{yue2019domain} during training. We surpass~\cite{yue2019domain} on both Cityscapes and IDD, respectively. We compare our method with two previous methods on MTDA. Since the previous works on MTDA only focus on the classification task, we carefully implement these methods in semantic segmentation with the same network. Compared to "MTDA-ITA", our method achieves significantly better performance on both domains. "MT-MTDA" is the method that adopts multiple teachers to alternatively teach a student in an offline knowledge distillation manner. However, the method also not consider to explore the information from different target domains. Our method achieves better performance than~\cite{nguyen2020unsupervised} on both Cityscapes and IDD.

\begin{table}[t]
	\footnotesize
	\setlength\tabcolsep{12pt}
	\caption{Results of adapting GTA5 to different target domains with ResNet-101 as backbone. "C", "I" and "M" represent "Cityscapes", "IDD" and "Mapillary", respectively.}
	\begin{center}
		\begin{tabular}{C{0.85cm} | C{0.16cm} C{0.16cm} C{0.16cm}| C{0.3cm}  C{0.3cm} C{0.3cm} }
			\toprule
			\multirow{2}{*}{Method} &\multicolumn{3}{c|}{Target} &\multicolumn{3}{c}{mIoU}\\
			&C &I &M  &C  &I &M\\
			\midrule
			\multirow{3}{*}{STDA}&\cmark & & &43.3 &- &- \\
			& &\cmark & &- &43.6 &- \\
			& & &\cmark &- &- &45.8 \\
			\midrule
			\multirow{4}{*}{MTDA}&\cmark &\cmark & &45.0 &46.0 &- \\
			&\cmark  & &\cmark  &45.1 &- &48.8 \\
			& &\cmark  &\cmark &- &44.5 &46.4 \\
			&\cmark &\cmark &\cmark &46.7 &47.0 &49.9 \\
			\bottomrule
		\end{tabular}
		\label{table:MTDA_syn}
	\end{center}
	\vspace{-5mm}
\end{table}

\subsection{Ablation Study}
In this section, we evaluate each component in the proposed CCL framework by conducting ablation studies on GTA5 to Cityscapes and IDD task with ResNet-101 as backbone. Results are shown in Table~\ref{table:ablation_CCL}.

We conduct a set of ablation study to examine the role of different components of the proposed method. A baseline (Model 1) here is designed as a method of directly applying adversarial loss to both target domains, \textit{i.e.,} $\lambda_{cl}=\lambda_{okd}=\lambda_{wr}=0$. When online knowledge distillation loss $\lambda_{okd}$ is switched on, Model 2 gains +1.0\% mIoU improvement on IDD but suffers from 0.5\% mIoU drops on Cityscapes. That could be explained by the confusion caused by the domain shift with expert models. When the weight regularization loss $\lambda_{wr}$ is switched on, Model 3 gains evident improvement of +0.8\% and +0.6\% mIoU than the baseline on Cityscapes and IDD. Using $\lambda_{okd}$ and $\lambda_{wr}$ simultaneously improve the Model 1 by 1.7\% and 1.8\% mIoU on Cityscapes and IDD, and also outperforms "Individual Model" in both target domains. Consistent improvement over Model 2, Model 3 and Model 4 is gained when collaborative consistency learning is employed. Specifically, Model 7 gains evident 1.0\% and 1.3\% improvement from Model 4 on Cityscapes and IDD, simultaneously.

\subsection{Generalization to Different Datasets}
\textbf{Synthetic-to-real MTDA.}
Here, we conduct a set of experiments with different target domains. We consider the task of STDA as our baseline, that includes: (1) GTA5 to Cityscapes, (2) GTA5 to IDD and (3) GTA5 to Mapillary. Each STDA model is trained on the corresponding target domain, individually. In Table~\ref{table:MTDA_syn}, three STDA baselines with three individually trained models achieve 43.3\%, 43.6\% and 45.8\% mIoU on Cityscapes, IDD and Mapillary, respectively. It can also be extended to adaptation to all these three datasets. Experiment results show that our method with a single model consistently works better than the STDA baseline, which is individually trained on the corresponding target domains. Our method using a single model consistently works better than the STDA baseline on the corresponding target domains.

\textbf{Real-to-real MTDA.}
In Table~\ref{table:MTDA_real}, we also conduct a domain experiment from real-world datasets to real-world datasets. Here one of the Cityscapes, IDD and Mapillary is adopted as the source domain and the rest two are taken as the two target domains. Experimental results show that the proposed method not only works well on syn-to-real adaptation but also does a good job on the case of real-to-real.

\begin{table}[t]
	\footnotesize
	\setlength\tabcolsep{12pt}
	\caption{Results for real-to-real MTDA experiments. }
	\vspace{2mm}
	\begin{center}
		\begin{tabular}{C{0.85cm} | C{0.16cm} C{0.16cm} C{0.16cm}| C{0.3cm}  C{0.3cm} C{0.3cm} }
			\toprule
			\multirow{2}{*}{Souce} &\multicolumn{3}{c|}{Target} &\multicolumn{3}{c}{mIoU}\\
			&C &I &M  &C  &I &M\\
			\midrule
			\multirow{3}{*}{C}& &\cmark &  &- &51.4 &- \\
			& & &\cmark &- &- &49.6 \\
			& &\cmark &\cmark &- &53.6  &51.4 \\
			\midrule
			\multirow{3}{*}{I}&\cmark & &  &46.5 &- &- \\
			& & &\cmark &- &- &49.0 \\
			&\cmark & &\cmark &46.8 &-  &49.8 \\
			\midrule
			\multirow{3}{*}{M}&\cmark & & &57.9 &- &- \\
			& &\cmark & &- &52.3 &- \\
			&\cmark &\cmark & &58.5 &54.1  &- \\
			\bottomrule
		\end{tabular}
		\label{table:MTDA_real}
	\end{center}
	\vspace{-5mm}
\end{table}

\section{Conclusion}
\label{conclusion}
In this work, we propose a novel collaborative consistency learning framework to achieve multi-target domain adaptation. The key idea is to first train a strong expert model for each target domain by simultaneously imposing consistency constraint among prediction from multiple expert models. They are further used as multiple teachers to collaboratively teach a student model in an online fashion such that a single model is able to work well across multiple target domains. Extensive experiments show that our method not only produces a single model that works well on multiple target domains but also achieves favorably performance against domain-specialized UDA methods on each domain.

	{\small
		\bibliographystyle{ieee_fullname}
          \bibliography{egbib}

\begin{thebibliography}{10}\itemsep=-1pt

\bibitem{balaji2018metareg}
Yogesh Balaji, Swami Sankaranarayanan, and Rama Chellappa.
\newblock Metareg: Towards domain generalization using meta-regularization.
\newblock In {\em NeurIPS}, 2018.

\bibitem{bottou2010large}
L{\'e}on Bottou.
\newblock Large-scale machine learning with stochastic gradient descent.
\newblock In {\em Proceedings of COMPSTAT'2010}. 2010.

\bibitem{chang2019all}
Wei-Lun Chang, Hui-Po Wang, Wen-Hsiao Peng, and Wei-Chen Chiu.
\newblock All about structure: Adapting structural information across domains
  for boosting semantic segmentation.
\newblock In {\em CVPR}, 2019.

\bibitem{chen2019progressive}
Chaoqi Chen, Weiping Xie, Wenbing Huang, Yu Rong, Xinghao Ding, Yue Huang,
  Tingyang Xu, and Junzhou Huang.
\newblock Progressive feature alignment for unsupervised domain adaptation.
\newblock In {\em CVPR}, 2019.

\bibitem{chen2017deeplab}
Liang-Chieh Chen, George Papandreou, Iasonas Kokkinos, Kevin Murphy, and Alan~L
  Yuille.
\newblock Deeplab: Semantic image segmentation with deep convolutional nets,
  atrous convolution, and fully connected crfs.
\newblock {\em TPAMI}, 40(4):834--848, 2017.

\bibitem{chen2017rethinking}
Liang-Chieh Chen, George Papandreou, Florian Schroff, and Hartwig Adam.
\newblock Rethinking atrous convolution for semantic image segmentation.
\newblock {\em CoRR}, abs/1706.05587, 2017.

\bibitem{chen2018encoder}
Liang-Chieh Chen, Yukun Zhu, George Papandreou, Florian Schroff, and Hartwig
  Adam.
\newblock Encoder-decoder with atrous separable convolution for semantic image
  segmentation.
\newblock In {\em ECCV}, 2018.

\bibitem{chen2019domain}
Minghao Chen, Hongyang Xue, and Deng Cai.
\newblock Domain adaptation for semantic segmentation with maximum squares
  loss.
\newblock In {\em ICCV}, 2019.

\bibitem{chen2018road}
Yuhua Chen, Wen Li, and Luc Van~Gool.
\newblock Road: Reality oriented adaptation for semantic segmentation of urban
  scenes.
\newblock In {\em CVPR}, 2018.

\bibitem{cordts2016cityscapes}
Marius Cordts, Mohamed Omran, Sebastian Ramos, Timo Rehfeld, Markus Enzweiler,
  Rodrigo Benenson, Uwe Franke, Stefan Roth, and Bernt Schiele.
\newblock The cityscapes dataset for semantic urban scene understanding.
\newblock In {\em CVPR}, 2016.

\bibitem{deng2009imagenet}
Jia Deng, Wei Dong, Richard Socher, Li-Jia Li, Kai Li, and Li Fei-Fei.
\newblock Imagenet: A large-scale hierarchical image database.
\newblock In {\em CVPR}, 2009.

\bibitem{dou2019domain}
Qi Dou, Daniel Coelho~de Castro, Konstantinos Kamnitsas, and Ben Glocker.
\newblock Domain generalization via model-agnostic learning of semantic
  features.
\newblock In {\em NeurIPS}, 2019.

\bibitem{furlanello2018born}
Tommaso Furlanello, Zachary~C Lipton, Michael Tschannen, Laurent Itti, and
  Anima Anandkumar.
\newblock Born again neural networks.
\newblock In {\em ICML}. 2018.

\bibitem{gholami2020unsupervised}
Behnam Gholami, Pritish Sahu, Ognjen Rudovic, Konstantinos Bousmalis, and
  Vladimir Pavlovic.
\newblock Unsupervised multi-target domain adaptation: An information theoretic
  approach.
\newblock {\em IEEE Transactions on Image Processing}, 2020.

\bibitem{he2016deep}
Kaiming He, Xiangyu Zhang, Shaoqing Ren, and Jian Sun.
\newblock Deep residual learning for image recognition.
\newblock In {\em CVPR}, 2016.

\bibitem{heo2019comprehensive}
Byeongho Heo, Jeesoo Kim, Sangdoo Yun, Hyojin Park, Nojun Kwak, and Jin~Young
  Choi.
\newblock A comprehensive overhaul of feature distillation.
\newblock In {\em ICCV}, 2019.

\bibitem{hinton2015distilling}
Geoffrey Hinton, Oriol Vinyals, and Jeff Dean.
\newblock Distilling the knowledge in a neural network.
\newblock In {\em NeurIPSW}, 2015.

\bibitem{hoffman2016fcns}
Judy Hoffman, Dequan Wang, Fisher Yu, and Trevor Darrell.
\newblock Fcns in the wild: Pixel-level adversarial and constraint-based
  adaptation.
\newblock {\em CoRR}, abs/1612.02649, 2016.

\bibitem{huang2020contextual}
Jiaxing Huang, Shijian Lu, Dayan Guan, and Xiaobing Zhang.
\newblock Contextual-relation consistent domain adaptation for semantic
  segmentation.
\newblock {\em ECCV}, 2020.

\bibitem{huang2017arbitrary}
Xun Huang and Serge Belongie.
\newblock Arbitrary style transfer in real-time with adaptive instance
  normalization.
\newblock In {\em ICCV}, 2017.

\bibitem{isobe2020video2}
Takashi Isobe, Xu Jia, Shuhang Gu, Songjiang Li, Shengjin Wang, and Qi Tian.
\newblock Video super-resolution with recurrent structure-detail network.
\newblock In {\em ECCV}, 2020.

\bibitem{isobe2020video}
Takashi Isobe, Songjiang Li, Xu Jia, Shanxin Yuan, Gregory Slabaugh, Chunjing
  Xu, Ya-Li Li, Shengjin Wang, and Qi Tian.
\newblock Video super-resolution with temporal group attention.
\newblock In {\em CVPR}, 2020.

\bibitem{isobe2020revisiting}
Takashi Isobe, Fang Zhu, and Shengjin Wang.
\newblock Revisiting temporal modeling for video super-resolution.
\newblock In {\em BMVC}, 2020.

\bibitem{kang2020towards}
Minsoo Kang, Jonghwan Mun, and Bohyung Han.
\newblock Towards oracle knowledge distillation with neural architecture
  search.
\newblock In {\em AAAI}, 2020.

\bibitem{khosla2012undoing}
Aditya Khosla, Tinghui Zhou, Tomasz Malisiewicz, Alexei~A Efros, and Antonio
  Torralba.
\newblock Undoing the damage of dataset bias.
\newblock In {\em ECCV}, 2012.

\bibitem{kim2020learning}
Myeongjin Kim and Hyeran Byun.
\newblock Learning texture invariant representation for domain adaptation of
  semantic segmentation.
\newblock In {\em CVPR}, 2020.

\bibitem{kingma2014adam}
Diederik~P Kingma and Jimmy Ba.
\newblock Adam: A method for stochastic optimization.
\newblock In {\em ICLR}, 2015.

\bibitem{lee2019sliced}
Chen-Yu Lee, Tanmay Batra, Mohammad~Haris Baig, and Daniel Ulbricht.
\newblock Sliced wasserstein discrepancy for unsupervised domain adaptation.
\newblock In {\em CVPR}, 2019.

\bibitem{li2020content}
Guangrui Li, Guoliang Kang, Wu Liu, Yunchao Wei, and Yi Yang.
\newblock Content-consistent matching for domain adaptive semantic
  segmentation.
\newblock In {\em ECCV}, 2020.

\bibitem{li2018deep}
Ya Li, Xinmei Tian, Mingming Gong, Yajing Liu, Tongliang Liu, Kun Zhang, and
  Dacheng Tao.
\newblock Deep domain generalization via conditional invariant adversarial
  networks.
\newblock In {\em ECCV}, 2018.

\bibitem{li2019bidirectional}
Yunsheng Li, Lu Yuan, and Nuno Vasconcelos.
\newblock Bidirectional learning for domain adaptation of semantic
  segmentation.
\newblock In {\em CVPR}, 2019.

\bibitem{lian2019constructing}
Qing Lian, Fengmao Lv, Lixin Duan, and Boqing Gong.
\newblock Constructing self-motivated pyramid curriculums for cross-domain
  semantic segmentation: A non-adversarial approach.
\newblock In {\em ICCV}, 2019.

\bibitem{long2015fully}
Jonathan Long, Evan Shelhamer, and Trevor Darrell.
\newblock Fully convolutional networks for semantic segmentation.
\newblock {\em TPAMI}, 39(4):640--651, 2017.

\bibitem{luo2019significance}
Yawei Luo, Ping Liu, Tao Guan, Junqing Yu, and Yi Yang.
\newblock Significance-aware information bottleneck for domain adaptive
  semantic segmentation.
\newblock In {\em ICCV}, 2019.

\bibitem{luo2019taking}
Yawei Luo, Liang Zheng, Tao Guan, Junqing Yu, and Yi Yang.
\newblock Taking a closer look at domain shift: Category-level adversaries for
  semantics consistent domain adaptation.
\newblock In {\em CVPR}, 2019.

\bibitem{lv2020cross}
Fengmao Lv, Tao Liang, Xiang Chen, and Guosheng Lin.
\newblock Cross-domain semantic segmentation via domain-invariant interactive
  relation transfer.
\newblock In {\em CVPR}, 2020.

\bibitem{mei2020instance}
Ke Mei, Chuang Zhu, Jiaqi Zou, and Shanghang Zhang.
\newblock Instance adaptive self-training for unsupervised domain adaptation.
\newblock In {\em ECCV}, 2020.

\bibitem{mirzadeh2020improved}
Seyed~Iman Mirzadeh, Mehrdad Farajtabar, Ang Li, Nir Levine, Akihiro Matsukawa,
  and Hassan Ghasemzadeh.
\newblock Improved knowledge distillation via teacher assistant.
\newblock In {\em AAAI}, 2020.

\bibitem{neuhold2017mapillary}
Gerhard Neuhold, Tobias Ollmann, Samuel Rota~Bulo, and Peter Kontschieder.
\newblock The mapillary vistas dataset for semantic understanding of street
  scenes.
\newblock In {\em ICCV}, 2017.

\bibitem{nguyen2020unsupervised}
Le~Thanh Nguyen-Meidine, Madhu Kiran, Jose Dolz, Eric Granger, Atif Bela, and
  Louis-Antoine Blais-Morin.
\newblock Unsupervised multi-target domain adaptation through knowledge
  distillation.
\newblock {\em CoRR}, abs/2007.07077, 2020.

\bibitem{noroozi2018boosting}
Mehdi Noroozi, Ananth Vinjimoor, Paolo Favaro, and Hamed Pirsiavash.
\newblock Boosting self-supervised learning via knowledge transfer.
\newblock In {\em CVPR}, 2018.

\bibitem{pan2020unsupervised}
Fei Pan, Inkyu Shin, Francois Rameau, Seokju Lee, and In~So Kweon.
\newblock Unsupervised intra-domain adaptation for semantic segmentation
  through self-supervision.
\newblock In {\em CVPR}, 2020.

\bibitem{reinhard2001color}
Erik Reinhard, Michael Adhikhmin, Bruce Gooch, and Peter Shirley.
\newblock Color transfer between images.
\newblock {\em IEEE Computer graphics and applications}, 21(5):34--41, 2001.

\bibitem{richter2016playing}
Stephan~R Richter, Vibhav Vineet, Stefan Roth, and Vladlen Koltun.
\newblock Playing for data: Ground truth from computer games.
\newblock In {\em ECCV}, 2016.

\bibitem{ros2016synthia}
German Ros, Laura Sellart, Joanna Materzynska, David Vazquez, and Antonio~M
  Lopez.
\newblock The synthia dataset: A large collection of synthetic images for
  semantic segmentation of urban scenes.
\newblock In {\em CVPR}, 2016.

\bibitem{simonyan2014very}
Karen Simonyan and Andrew Zisserman.
\newblock Very deep convolutional networks for large-scale image recognition.
\newblock 2015.

\bibitem{tsai2018learning}
Yi-Hsuan Tsai, Wei-Chih Hung, Samuel Schulter, Kihyuk Sohn, Ming-Hsuan Yang,
  and Manmohan Chandraker.
\newblock Learning to adapt structured output space for semantic segmentation.
\newblock In {\em CVPR}, 2018.

\bibitem{tsai2019domain}
Yi-Hsuan Tsai, Kihyuk Sohn, Samuel Schulter, and Manmohan Chandraker.
\newblock Domain adaptation for structured output via discriminative patch
  representations.
\newblock In {\em ICCV}, 2019.

\bibitem{varma2019idd}
Girish Varma, Anbumani Subramanian, Anoop Namboodiri, Manmohan Chandraker, and
  CV Jawahar.
\newblock Idd: A dataset for exploring problems of autonomous navigation in
  unconstrained environments.
\newblock In {\em WACV}, 2019.

\bibitem{vu2019advent}
Tuan-Hung Vu, Himalaya Jain, Maxime Bucher, Matthieu Cord, and Patrick
  P{\'e}rez.
\newblock Advent: Adversarial entropy minimization for domain adaptation in
  semantic segmentation.
\newblock In {\em CVPR}, 2019.

\bibitem{vu2019dada}
Tuan-Hung Vu, Himalaya Jain, Maxime Bucher, Matthieu Cord, and Patrick
  P{\'e}rez.
\newblock Dada: Depth-aware domain adaptation in semantic segmentation.
\newblock In {\em ICCV}, 2019.

\bibitem{wang2020differential}
Zhonghao Wang, Mo Yu, Yunchao Wei, Rogerio Feris, Jinjun Xiong, Wen-mei Hwu,
  Thomas~S Huang, and Honghui Shi.
\newblock Differential treatment for stuff and things: A simple unsupervised
  domain adaptation method for semantic segmentation.
\newblock In {\em CVPR}, 2020.

\bibitem{yang2020label}
Jinyu Yang, Weizhi An, Sheng Wang, Xinliang Zhu, Chaochao Yan, and Junzhou
  Huang.
\newblock Label-driven reconstruction for domain adaptation in semantic
  segmentation.
\newblock In {\em ECCV}, 2020.

\bibitem{yang2020fda}
Yanchao Yang and Stefano Soatto.
\newblock Fda: Fourier domain adaptation for semantic segmentation.
\newblock In {\em CVPR}, 2020.

\bibitem{yu2020context}
Changqian Yu, Jingbo Wang, Changxin Gao, Gang Yu, Chunhua Shen, and Nong Sang.
\newblock Context prior for scene segmentation.
\newblock In {\em CVPR}, 2020.

\bibitem{yu2018multi}
Huanhuan Yu, Menglei Hu, and Songcan Chen.
\newblock Multi-target unsupervised domain adaptation without exactly shared
  categories.
\newblock {\em CoRR}, abs/1809.00852, 2018.

\bibitem{yue2019domain}
Xiangyu Yue, Yang Zhang, Sicheng Zhao, Alberto Sangiovanni-Vincentelli, Kurt
  Keutzer, and Boqing Gong.
\newblock Domain randomization and pyramid consistency: Simulation-to-real
  generalization without accessing target domain data.
\newblock In {\em ICCV}, 2019.

\bibitem{zhang2020generalizable}
Jian Zhang, Lei Qi, Yinghuan Shi, and Yang Gao.
\newblock Generalizable semantic segmentation via model-agnostic learning and
  target-specific normalization.
\newblock {\em CoRR}, abs/2003.12296, 2020.

\bibitem{zhang2019curriculum}
Yang Zhang, Philip David, Hassan Foroosh, and Boqing Gong.
\newblock A curriculum domain adaptation approach to the semantic segmentation
  of urban scenes.
\newblock {\em TPAMI}, 42(8):1823--1841, 2020.

\bibitem{zhang2018deep}
Ying Zhang, Tao Xiang, Timothy~M Hospedales, and Huchuan Lu.
\newblock Deep mutual learning.
\newblock In {\em CVPR}, 2018.

\bibitem{zhao2019multi}
Sicheng Zhao, Bo Li, Xiangyu Yue, Yang Gu, Pengfei Xu, Runbo Hu, Hua Chai, and
  Kurt Keutzer.
\newblock Multi-source domain adaptation for semantic segmentation.
\newblock In {\em NeurIPS}, 2019.

\bibitem{zhu2017unpaired}
Jun-Yan Zhu, Taesung Park, Phillip Isola, and Alexei~A Efros.
\newblock Unpaired image-to-image translation using cycle-consistent
  adversarial networks.
\newblock In {\em ICCV}, 2017.

\bibitem{zhu2019asymmetric}
Zhen Zhu, Mengde Xu, Song Bai, Tengteng Huang, and Xiang Bai.
\newblock Asymmetric non-local neural networks for semantic segmentation.
\newblock In {\em ICCV}, 2019.

\bibitem{zou2018unsupervised}
Yang Zou, Zhiding Yu, BVK Vijaya~Kumar, and Jinsong Wang.
\newblock Unsupervised domain adaptation for semantic segmentation via
  class-balanced self-training.
\newblock In {\em ECCV}, 2018.

\end{thebibliography}
	}
	
\end{document}